\documentclass{article}

\usepackage{spconf,amsmath,graphicx,hyperref}
\usepackage{glossaries}
\usepackage{amsfonts}
\usepackage{caption}
\usepackage{subfloat}
\usepackage{subcaption}
\usepackage{booktabs,siunitx,tabularx, stfloats}
\usepackage{xcolor}
\usepackage{flushend}

\newacronym{gan}{GAN}{Generative Adversarial Network}
\newacronym{da}{DA}{Data Augmentation}
\newacronym{nn}{NN}{Neural Network}
\newacronym{cnn}{CNN}{Convolutional Neural Network}
\newacronym{cv}{CV}{Computer Vision}
\newacronym{sd}{SD}{Stable Diffusion}
\newacronym{sde}{SDE}{Stochastic Differential Equation}
\newacronym{ldm}{LDM}{Latent Diffusion Model}
\newacronym{dm}{DM}{Diffusion Model}
\newacronym{csam}{CSAM}{Child Sexual Abuse Material}
\newacronym{auc}{AUC}{Area Under the Curve}
\newacronym{ba}{B-ACC}{Balanced Accuracy}
\newacronym{ba0}{$\textrm{B-ACC}_{@ \textrm{thr}=0}$}{Balanced Accuracy at threshold 0}
\newacronym{tpr0}{$\textrm{TPR}_{@\textrm{th}}$}{True Positive Rate at fixed threshold}
\newacronym{fpr0}{$\textrm{FPR}_{@ \textrm{thr}=0}$}{False Positive Rate at threshold 0}
\newacronym{lpip}{LPIPS}{Learned Perceptual Image Patch Similarity}
\newacronym{psnr}{PSNR}{Peak Signal-to-Noise Ratio}
\newacronym{ssim}{SSIM}{Structural Similarity Index Measure}

\def\M{{\mathbf M}}
\def\H{{\mathbf H}}

\def\Fx{{\mathcal{F}_x}}
\def\Fy{{\mathcal{F}_y}}

\title{Beyond Spectral Peaks: \\ Interpreting the Cues Behind Synthetic Image Detection}
\name{Sara Mandelli\(^1\), Diego Vila-Portela\(^2\), David Vázquez-Padín\(^2\), Paolo Bestagini\(^1\), Fernando Pérez-González\(^2\)}
\address{\(^1\)Dipartimento di Elettronica, Informazione e Bioingegneria, Politecnico di Milano - Milan, Italy \\
\(^2\)atlanTTic Research Center, University of Vigo, E.E. de Telecomunicación - Vigo, Spain\\
\thanks{This work was supported by the FOSTERER project, funded by the Italian Ministry of Education, University, and Research within the PRIN 2022 program.
This work was also partially supported by the European Union - Next Generation EU under the Italian National Recovery and Resilience Plan (NRRP), Mission 4, Component 2, Investment 1.3: CUP D43C22003080001, partnership on ``Telecommunications of the Future'' (PE00000001 - program ``RESTART''); CUP D43C22003050001, partnership on ``SEcurity and RIghts in the CyberSpace'' (PE00000014 - program ``FF4ALL-SERICS''). This work was also supported by Xunta de Galicia and the European Regional Development Fund under Project ED431C 2025/41.}}

\begin{document}
\ninept

\maketitle
\begin{abstract}
Over the years, the forensics community has proposed several deep learning-based detectors to mitigate the risks of generative AI.
Recently, frequency-domain artifacts (particularly periodic peaks in the magnitude spectrum), have received significant attention, as they have been often considered a strong indicator of synthetic image generation.
However, state-of-the-art detectors are typically used as black-boxes, and it still remains unclear whether they truly rely on these peaks. This limits their interpretability and trust.

In this work, we conduct a systematic study to address this question. We propose a strategy to remove spectral peaks from images and analyze the impact of this operation on several detectors. In addition, we introduce a simple linear detector that relies exclusively on frequency peaks, providing a fully interpretable baseline free from the confounding influence of deep learning.
Our findings reveal that most detectors are not fundamentally dependent on spectral peaks, challenging a widespread assumption in the field and paving the way for more transparent and reliable forensic tools.
\end{abstract}

\section{Introduction}
\label{sec:intro}

The advent of generative AI has fundamentally changed the way synthetic content is produced, making it possible for virtually anyone to generate high-quality media without advanced technical knowledge.
Although such technologies hold promise for creative industries and data enhancement~\cite{yang2024pixel}, they have also raised serious concerns about privacy, security, and the dissemination of misinformation~\cite{amerini2025deepfake}.
The proliferation of deepfake generation techniques amplifies the potential for malicious use, ranging from identity theft and non-consensual pornography to political disinformation and fraud. 

To counteract the spreading of malicious deepfakes, a wide range of forensic detectors has been proposed in recent years, almost all of which rely on deep learning~\cite{corvi2023detection, guillaro2025bias, cozzolino2024raising, sha2023fake, mandelli2024synthetic, baraldi2024contrasting, koutlis24}. However, since these models often operate as ``black boxes'', one of the main challenges lies in the interpretability of their outcomes. In particular, it remains hard to understand which generative artifacts the detectors exploit in order to distinguish synthetic content from authentic data.

Interpreting the output of a forensic detector remains a highly complex task, as deep learning systems often lack transparency in their decision-making process. 
A critical concern is that a detector may inherit biases from its training dataset, leading it to base its predictions on spurious correlations rather than genuine generative artifacts. For example, differences in compression levels, semantic content, or other dataset-specific characteristics between real and synthetic images can unintentionally guide the detector’s classification~\cite{guillaro2025bias}. This raises important questions regarding the reliability, generalization, and fairness of current deep learning–based detection methods.
Moreover, even in the absence of dataset biases, when a detector genuinely captures characteristic traces of the generation process, explaining which specific cues are being leveraged remains highly challenging. It is also likely that different detectors rely on distinct types of generative traces, with some focusing on certain patterns and others exploiting entirely different signals. 

Over the years, the forensic community has reported frequency domain artifacts as an important trace that allows to tell real and synthetic images apart~\cite{guillaro2025bias, mandelli2024synthetic, Corvi2023Intriguing}. 
In particular, significant attention has been devoted to energy peaks occurring in the magnitude spectrum,
which are often considered a strong indicator of synthetic image generation.
However, the vast majority of prior research has limited to describing these spectral artifacts without explicitly leveraging them as discriminative cues for detection. This limitation is particularly evident in deep learning–based detectors, where interpreting the origin of a specific prediction remains challenging.

In this work, we take a step toward shedding light on the actual interpretability of synthetic image detectors. A central question we address is whether deep learning–based detectors truly rely on spectral peaks introduced by synthetic generation, or if these artifacts play only a marginal role in the decision process. To investigate this, we conduct a systematic analysis of several state-of-the-art detectors and design experiments aimed at disentangling their reliance on such frequency-domain cues. 
Specifically, we design a strategy to remove periodic peaks from the frequency spectrum of images and we assess how this operation affects different detectors. In addition, we introduce a very simple detector that relies exclusively on frequency peaks, without any data-driven component. This experiment enables a clearer interpretation of the results, free from the possible confounding effects of deep learning’s black-box nature.

Our results suggest that, for most detectors, the presence of spectral peaks does not constitute a fundamental artifact for detection, challenging a common assumption in the field and opening the way to a deeper understanding of what features these models exploit.

\section{Frequency-domain generation artifacts}
\label{sec:background}

In the forensic community, it is widely recognized that synthetic image generation techniques introduce distinctive traces in their produced content~\cite{mandelli2024synthetic, Corvi2023Intriguing, bammey2023synthbuster}. 
All generated images exhibit such artifacts, whether produced from a text prompt or via the ``img2img'' modality (i.e., where a new image is synthesized from an existing one).
Recent studies have further shown that even images simply passed through a generative model’s autoencoder (without any diffusion step) display artifacts similar to those found in fully synthetic generation~\cite{mandelli2024synthetic}. 
These images have been referred to as ``laundered'', since their semantic content is almost entirely preserved, with only minor imperceptible alterations, while still retaining synthetic-like traces in the frequency domain.

The generation artifacts often manifest as pronounced peaks in the Fourier spectrum of noise residuals extracted from synthetic images, typically appearing at components with periods of $4$, $8$ or $16$ samples in both directions~\cite{bammey2023synthbuster}. Together with these, artifacts may also appear as recurring structures like rings, ovals, or circular patterns in the Fourier domain~\cite{coccomini2024deepfake}. Such artifacts are generally attributed to the upsampling operators employed during the decoding stage~\cite{durall2020watch}, although they may also arise from the characteristics of the training dataset used for a specific generator~\cite{Corvi2023Intriguing}.

\begin{figure}[t]
        \centering
        \includegraphics[width=\columnwidth]{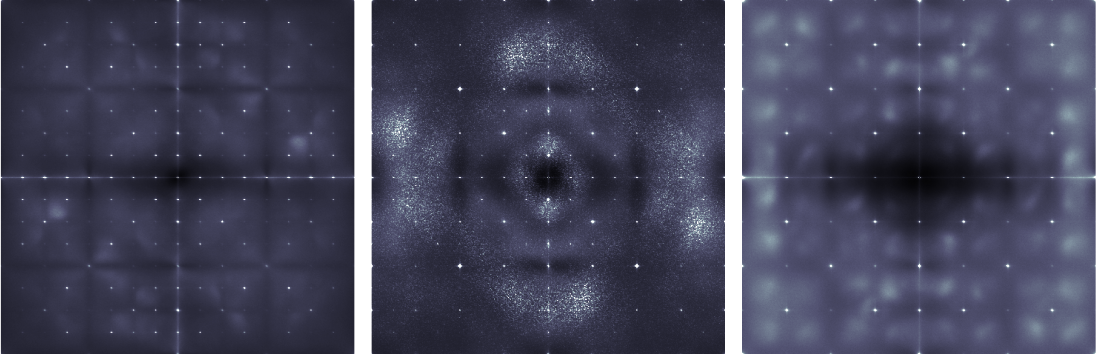}
        \vspace{-15pt}
        \caption{Fourier transform analysis of synthetic images generated, respectively from left to right by \gls{sd}3.5~\cite{stable_diffusion_35}, Flux 1.1Pro~\cite{fluxpro} and DALL$\cdot$E\,3~\cite{dalle3}. 
        }
        \label{fig:frequency_artifacts_wild}
        \vspace{-5pt}
\end{figure}

Artifact analysis is typically performed by subjecting the images to a high-pass filtering process, frequently implemented through a neural network. For instance, a very common choice is to adopt the DnCNN architecture~\cite{zhang2017beyond} as a denoiser to reveal peaks and other spectral irregularities~\cite{mandelli2024synthetic, Corvi2023Intriguing}. 
To illustrate the frequency artifacts commonly observed in synthetic images, Fig.~\ref{fig:frequency_artifacts_wild} reports the average power spectra computed from synthetic images of different generators included in the recently released Wild dataset~\cite{bongini2025wild}. 
Before averaging, all images have been processed with the DnCNN-based denoiser proposed in~\cite{Corvi2023Intriguing}.
The spectra reveal a strong presence of peaks with different periodicity across all generators; moreover, each generator exhibits distinct spectral traces that could potentially allow for the unique characterization of its images.

Several recent studies have shown that synthetic and real images can be distinguished by analyzing their frequency spectra~\cite{guillaro2025bias, mandelli2024synthetic, Corvi2023Intriguing, bammey2023synthbuster, cannas2024jpeg}. 
However, most prior research has limited to showing these spectral discrepancies rather than explicitly using them as discriminative detection traces. 
To our knowledge, only few studies have directly targeted frequency-domain mismatches between real and synthetic data~\cite{bammey2023synthbuster, zhang2019detecting, dzanic2020fourier, frank2020leveraging}. 
Moreover, all these works focused on artifacts from relatively old generators, with only SynthBuster~\cite{bammey2023synthbuster} extending the investigation to diffusion models.
\vspace{-6pt}
\section{Proposed experimental analysis}
\label{sec:proposed_analysis}

To take a step toward improving the interpretability of detectors, we focus on the Fourier spectrum of the tested images. 
Specifically, we design two experiments:
(i) peak removal from synthetic images; (ii) peak removal from laundered images.

In a nutshell, we apply a binary mask operator to the entire spectrum of the images (phase included) for removing the energy of the spectral components around the peaks. 
Our goal is to assess how detectors' performance changes when these spectral modifications are applied. 
If detectors relied primarily on spectral peaks, we would expect their scores to vary significantly after peak removal. 
Conversely, if the spectral energy at peak positions were not a key factor for detection, the scores should remain largely unaffected. 



\textbf{Peaks removal from synthetic images. }
As shown in Section~\ref{sec:background}, the Fourier spectrum of each generator exhibits peaks at different positions. To suppress these peaks, we adopt a straightforward masking strategy in the Fourier domain, designed to remove frequency components lying on a $P \times P$ grid, where $P$ denotes the periodicity of the peaks to be eliminated.

Formally, we define two sets of normalized spatial frequencies along two dimensions, $\Fx =\{n/P, \; n \in \mathbb{N}\}$ and $\Fy =\{m/P, \; m \in \mathbb{N}\}$.
Then, we define a binary mask $\M(f_x, f_y)$ as
\begin{equation}
    \M(f_x, f_y) = \begin{cases}
    0,\quad \text{if $(f_x, f_y) \in \Fx \times \Fy \setminus \{(0, 0)\}$}\\
    1,\quad \text{elsewhere}
    \end{cases}.
    \label{eq:mask_def}
\end{equation}
To delete the peaks, we apply the mask to the full spectrum of each input image in a coefficient-wise fashion. 
However, we experimentally verified that simply deleting the periodic pattern at the exact peak positions is insufficient to fully suppress the spectral energy associated with the peaks. To address this, we apply a dilation operator with a disk-shaped structuring element,
thereby slightly enlarging the ``holes'' in their surrounding area. 
Note that we preserve the peak at frequency $(0,0)$ to avoid altering the low-pass behaviour of the image, i.e., its mean value and immediate surroundings. 
After applying the mask, we go back to the pixel domain, adjusting the output dynamics to match that of the input image, and we quantize it to $8$-bit\footnote{The peak removal code can be found at \href{https://github.com/polimi-ispl/beyond-spectral-peaks}{\url{https://github.com/polimi-ispl/beyond-spectral-peaks}}\label{footnote_ref}}.

\begin{figure*}[t]
        \includegraphics[width=\textwidth]{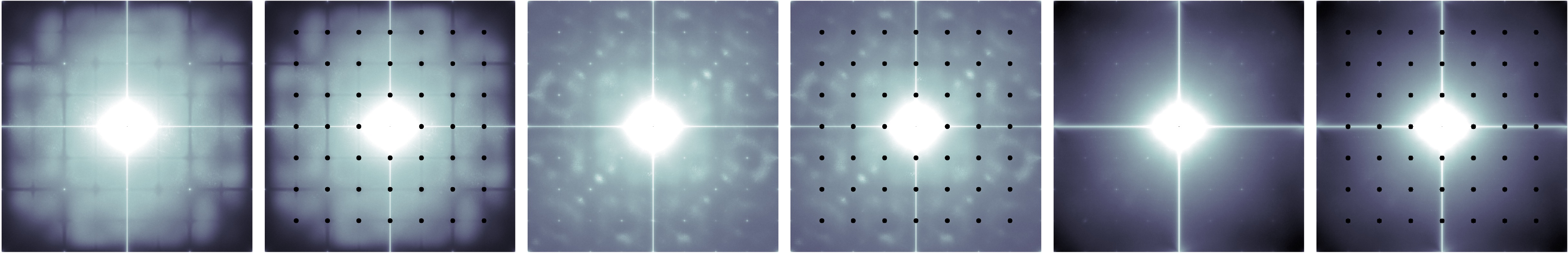}
        \vspace{-15pt}
        \caption{Average Fourier spectra (magnitude, in logarithmic scale) of synthetic images generated with Midjourney (first column), DALL$\cdot$E 3 (third) and \gls{sd}XL (fifth) before and after peak removal with periodicity $P = 8$. Best viewed in electronic format.}
        \label{fig:mask_removal_wild}
\vspace{-10pt}
\end{figure*}

Fig.~\ref{fig:mask_removal_wild} illustrates the effect of this procedure by reporting the average power spectra of $1000$ synthetic images generated with Midjourney~\cite{midjourney}, DALL$\cdot$E\,3~\cite{dalle3} and \gls{sd}XL~\cite{podell2023sdxl}, before and after peak removal with periodicity $P = 8$. To show these examples, we omit the denoising step to directly highlight the plain spectral magnitudes before and after the proposed modification.

\begin{figure}[t]
        \includegraphics[width=\columnwidth]{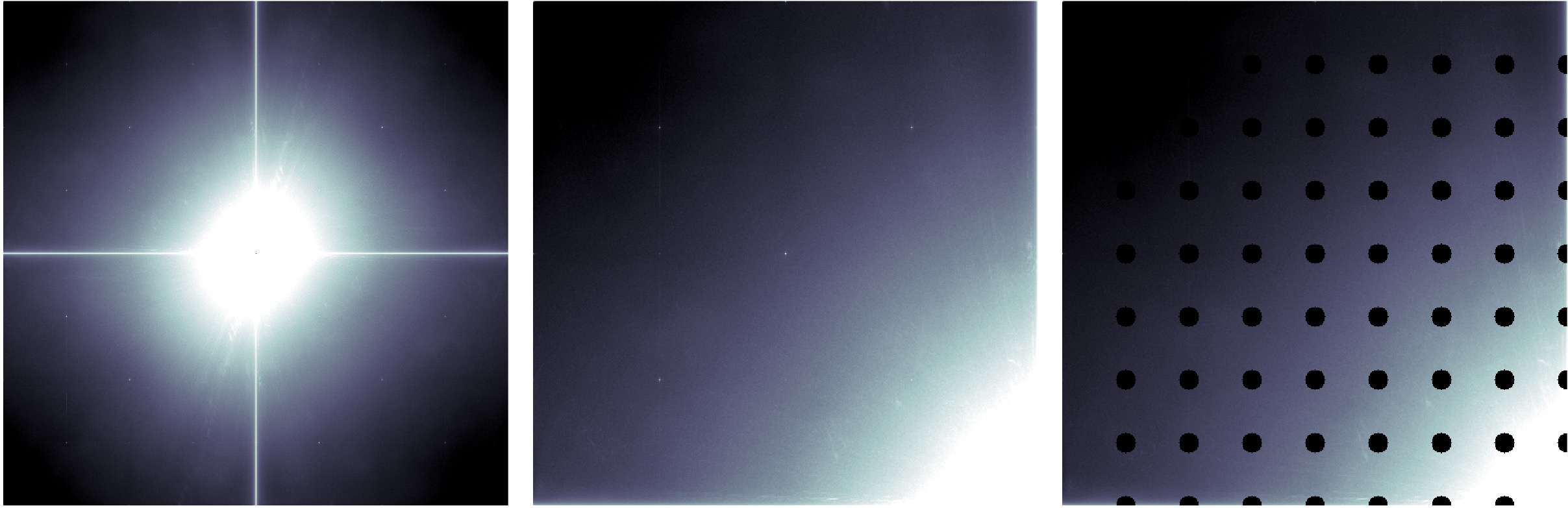}
        \vspace{-15pt}
        \caption{From left to right: average Fourier spectrum (magnitude, in logarithmic scale) of laundered versions of real images through \gls{sd}3.5; close-up of one quadrant; close-up of the peak-removed spectrum with periodicity $P = 16$. Best viewed in electronic format.}
        \label{fig:mask_removal_laundered}
        \vspace{-12pt}
\end{figure}

\textbf{Peaks removal from laundered images.}
Alongside synthetic images, we also evaluate detectors on laundered images, i.e., real images passed through an encoding–decoding chain to erase their original traces and simulate synthetic generation~\cite{cozzolino2024raising, mandelli2024synthetic}.
As a matter of fact, image laundering poses a concrete threat in forensic contexts, as it provides an effective way to conceal user traces and disguise content as if it were synthetically generated. 
Recently, it has been shown that even the most advanced detectors may be highly vulnerable to this manipulation, often misclassifying laundered images as synthetic~\cite{mandelli2024synthetic}. As potential consequence, sensitive or harmful material may be overlooked or not appropriately flagged, increasing the risk of its dissemination online. 

To experiment on these images, we apply the same peak-removal pipeline described previously for fully synthetic content. Fig.~\ref{fig:mask_removal_laundered} illustrates an example of average spectral magnitudes computed from laundered images produced using the autoencoder of \gls{sd}3.5~\cite{stable_diffusion_35}. As it can be inspected, the laundered spectrum contains less visible artifacts than that of fully synthetic images; nonetheless, the frequency peaks are still there (kindly refer to the close-up shown in Fig.~\ref{fig:mask_removal_laundered}).

\vspace{-5pt}

\section{Experimental Setup}
\label{sec:setup}

\textbf{Dataset. }Synthetic images have been selected from the recently released Wild dataset~\cite{bongini2025wild}, which was built using some of the most popular commercial and open-source generators currently available. Specifically, we consider its ``closed-set'' sub-dataset, containing $1000$ text-to-image samples for each of $10$ generators, with average resolution of $1024 \times 1024$ pixels. All images are provided in uncompressed format, with the exception of those produced by Adobe Firefly~\cite{firefly} and Freepik~\cite{freepik}, which are JPEG-compressed files. 

Real images have been selected from the Raise dataset~\cite{dang2015raise}; specifically, we randomly selected $1000$ images and cropped them to a standard size of $1024 \times 1024$ pixels from the top-left corner. Laundered versions of these images were then synthesized following the procedure described in~\cite{mandelli2024synthetic}, using recent open-source generators, namely \gls{sd}XL~\cite{podell2023sdxl}, \gls{sd}3.5~\cite{stable_diffusion_35} and Flux.1~\cite{flux2024}.

Importantly, all images used in our analysis are kept uncompressed. 
We do this on purpose, as it is well known that JPEG compression introduces a characteristic $8 \times 8$ grid pattern in the frequency domain, and this can obscure synthetic generation traces~\cite{bammey2023synthbuster}. 
Similar distortions could arise from resizing or rotation, as these operations typically leave interpolation artifacts in the spectrum. 
Since our goal is to investigate whether the removal of frequency peaks specific to synthetic generation affects forensic detectors, we restrict our analysis to uncompressed images not undergone any post-processing. For this reason, we also omit synthetic images of Adobe Firefly and Freepik from our analysis.

\textbf{Synthetic image detectors.} We evaluate several state-of-the-art detectors spanning from CNNs to transformer-based architectures, trained under different paradigms and on diverse datasets. Specifically, we consider the detectors proposed in~\cite{corvi2023detection, guillaro2025bias, cozzolino2024raising, sha2023fake, mandelli2024synthetic, baraldi2024contrasting, koutlis24}, all of them being publicly available, with both testing code and pretrained weights. For consistency, we run all experiments using the official implementations released by the original authors.

We initially considered also the SynthBuster detector~\cite{bammey2023synthbuster} which, as discussed in Section~\ref{sec:background}, relies on feeding a machine learning model with information about frequency peaks. 
However, 
it experimentally reported excessively high false alarms on original content, and, 
for this reason, we excluded it from our evaluation.

Nonetheless, inspired by the SynthBuster strategy, we also evaluate an extremely simple linear detector that focuses solely on the frequency peaks, without relying on any data-driven approach. This allows us to provide a clearer interpretation of the results, without the confounding influence of the black-box nature of deep learning solutions.
Our developed detector extracts a high-pass residue from each image by applying a Laplacian of Gaussian Kernel $\H$~\cite{marr1980theory}:
\begin{equation}
    [\H]_{n_1,n_2}=-\frac{1}{\pi\sigma^4}\bigg[ 1-\frac{n_1^2+n_2^2}{2\sigma^2} \bigg]e^{-\frac{n_1^2+n_2^2}{2\sigma^2}}, 
\end{equation}
with $\sigma = 0.7$ and $n_1, n_2 \in [-5, 5]$.
Then, the detector computes the magnitude spectrum of the residual and averages the frequency contributions lying on a $8 \times 8$ or on a $16 \times 16$ grid, obtaining a real score that can be used for classifying the images. 

\textbf{Evaluation metrics.} We recall that each deep learning-based detector outputs a score $s$ that is thresholded at $0$ to distinguish between real ($s \leq 0$) and synthetic ($s > 0$) images.
To quantify the effects of peaks removal, we evaluate the percentage of images that are classified as synthetic.
Since all images we are working with are either fully synthetic or laundered, we consider all of them as part of the ``positive'' set, exploiting the \gls{tpr0} as metrics for evaluating the detectors before and after peaks removal. This threshold is equal to $0$ for all deep learning-based detectors. 
For the linear detectors (i.e., that with grid $8$ and the one with grid $16$), we calibrated them such to obtain a $5\%$ of false alarms over an internal dataset of real images. 
The resulting threshold has been applied for evaluating them over the testing data.
\vspace{-10pt}

\section{Results}
\label{sec:results}

\begin{table*}[t]
\centering
\caption{\gls{tpr0} achieved over uncompressed images of the Wild dataset (without peaks removal). In bold, we highlight average results greater than $70\%$.
}
\vspace{-5pt}
\label{tab:wild_results}
\resizebox{.75\textwidth}{!}{
\begin{tabular}{@{}lcccccccccccc@{}}
\toprule
Detector & DALL$\cdot$E\,3 & Flux.1 & Flux 1.1Pro & Leonardo AI & Midjourney & SD3.5 & SDXL & Starry AI & Average \\
\toprule
\cite{corvi2023detection} & $1.000$ & $0.727$ & $0.932$ & $1.000$ & $0.841$ & $1.000$ & $0.269$ & $1.000$ & $\mathbf{0.846}$ \\
\cite{cozzolino2024raising} & $0.675$ & $0.131$ & $0.108$ & $0.040$ & $0.046$ & $0.337$ & $0.071$ & $0.257$ & $0.208$ \\
\cite{mandelli2024synthetic} & $1.000$ & $1.000$ & $1.000$ & $1.000$ & $0.984$ & $1.000$ & $1.000$ & $1.000$ & $\mathbf{0.998}$ \\
\cite{guillaro2025bias} & $1.000$ & $0.738$ & $0.975$ & $0.915$ & $0.632$ & $0.997$ & $0.866$ & $0.998$ & $\mathbf{0.890}$ \\
\cite{baraldi2024contrasting} & $0.935$ & $0.755$ & $0.593$ & $0.539$ & $0.395$ & $0.650$ & $0.236$ & $0.878$ & $0.623$ \\
\cite{sha2023fake} & $0.963$ & $0.965$ & $0.965$ & $0.986$ & $0.920$ & $0.950$ & $0.699$ & $0.932$ & $\mathbf{0.922}$ \\
\cite{koutlis24} & $0.997$ & $0.883$ & $0.993$ & $0.871$ & $0.638$ & $0.947$ & $0.946$ & $0.219$ & $\mathbf{0.812}$ \\
Linear-$8$ & $0.985$ & $0.644$ & $1.000$ & $1.000$ & $1.000$ & $0.594$ & $0.536$ & $0.304$ & $\mathbf{0.859}$ \\
Linear-$16$& $0.938$ & $0.516$ & $1.000$ & $1.000$ & $0.997$ & $0.974$ & $0.334$ & $0.233$ & $\mathbf{0.750}$ \\
\bottomrule
\end{tabular}
}
\vspace{-5pt}
\end{table*}

\begin{table*}[t!]
\centering
\caption{Relative difference of the \gls{tpr0} achieved with peaks removal at periodicity $8$ (i.e., $\mathrm{R_m}8$) and periodicity $16$ ($\mathrm{R_m}16$) with respect to standard conditions, over the Wild synthetic dataset. We do not report results for deep learning-based detectors in which the original \gls{tpr0} was below $70\%$. In bold, we highlight results with absolute value greater than $30\%$.}
\vspace{-5pt}
\label{tab:rel_diff_rm_wild_agg}
\resizebox{\textwidth}{!}{
\begin{tabular}{@{}lcccccccccccc@{}}
\toprule
Detector & DALL$\cdot$E\,3 & Flux.1 & Flux 1.1Pro & Leonardo AI & Midjourney & SD3.5 & SDXL & Starry AI & Average\\
\toprule
 & $\mathrm{R_m}8$/$\mathrm{R_m}16$ & $\mathrm{R_m}8$/$\mathrm{R_m}16$ & $\mathrm{R_m}8$/$\mathrm{R_m}16$ & $\mathrm{R_m}8$/$\mathrm{R_m}16$ & $\mathrm{R_m}8$/$\mathrm{R_m}16$ & $\mathrm{R_m}8$/$\mathrm{R_m}16$ & $\mathrm{R_m}8$/$\mathrm{R_m}16$ & $\mathrm{R_m}8$/$\mathrm{R_m}16$ &
 $\mathrm{R_m}8$/$\mathrm{R_m}16$\\
\midrule
\cite{corvi2023detection} & $-0.00$/$-0.10$ & $\mathbf{-0.98}$/$\mathbf{-0.92}$ & $\mathbf{-0.54}$/$\mathbf{-0.63}$ & $+0.00$/$+0.00$ & $\mathbf{-0.86}$/$\mathbf{-0.98}$ & $\mathbf{-0.80}$/$\mathbf{-0.89}$ & $--$/$--$ & $-0.00$/$-0.05$ & $\mathbf{-0.45}$/$\mathbf{-0.51}$ \\
\cite{mandelli2024synthetic} & $-0.00$/$-0.01$ & $-0.00$/$-0.00$ & $+0.00$/$-0.00$ & $+0.00$/$+0.00$ & $-0.09$/$\mathbf{-0.47}$ & $-0.00$/$-0.02$ & $+0.00$/$-0.01$ & $+0.00$/$+0.00$ & $-0.01$/$-0.06$ \\
\cite{guillaro2025bias} & $+0.00$/$+0.00$ & $\mathbf{-0.46}$/$\mathbf{-0.83}$ & $-0.05$/$-0.19$ & $-0.01$/$-0.08$ & $--$/$--$ & $-0.01$/$-0.14$ & $-0.24$/$\mathbf{-0.49}$ & $-0.01$/$-0.05$ & $-0.11$/$-0.26$ \\
\cite{sha2023fake} & $+0.01$/$-0.04$ & $-0.04$/$-0.14$ & $-0.01$/$-0.03$ & $+0.00$/$-0.02$ & $-0.02$/$-0.16$ & $-0.02$/$-0.10$ & $--$/$--$ & $-0.09$/$-0.15$ & $-0.02$/$-0.09$ \\
\cite{koutlis24} & $+0.00$/$-0.00$ & $+0.06$/$-0.02$ & $+0.00$/$+0.01$ & $+0.07$/$+0.06$ & $--$/$--$ & $+0.02$/$-0.01$ & $+0.00$/$-0.02$ & $--$/$--$ & $+0.02$/$+0.00$ \\
Linear-$8$& $\mathbf{-0.95}$/$\mathbf{-0.94}$ & $\mathbf{-0.99}$/$\mathbf{-0.99}$ & $-0.19$/$-0.17$ & $\mathbf{-0.52}$/$\mathbf{-0.51}$ & $\mathbf{-0.99}$/$\mathbf{-0.99}$ & $\mathbf{-1.00}$/$\mathbf{-1.00}$ & $\mathbf{-1.00}$/$\mathbf{-1.00}$ & $\mathbf{-1.00}$/$\mathbf{-1.00}$ & $\mathbf{-0.83}$/$\mathbf{-0.83}$ \\
Linear-$16$& $\mathbf{-0.90}$/$\mathbf{-0.99}$ & $\mathbf{-0.97}$/$\mathbf{-1.00}$ & $+0.00$/$\mathbf{-0.75}$ & $\mathbf{-0.58}$/$\mathbf{-0.99}$ & $\mathbf{-0.94}$/$\mathbf{-1.00}$ & $-0.150$/$\mathbf{-1.00}$ & $\mathbf{-1.00}$/$\mathbf{-1.00}$ & $\mathbf{-0.98}$/$\mathbf{-1.00}$ & $\mathbf{-0.69}$/$\mathbf{-0.97}$ \\
\bottomrule
\end{tabular}
}
\vspace{-10pt}
\end{table*}

\begin{table}[t]
\centering
\caption{Relative difference of the \gls{tpr0} achieved with peaks removal at periodicity $8$ and $16$ with respect to standard conditions, over laundered images. In bold, we highlight results with absolute value greater than $30\%$.}
\vspace{-5pt}
\label{tab:tpr_laundered_rel_diff}
\resizebox{.95\columnwidth}{!}{
\begin{tabular}{@{}lccccccc@{}}
\toprule
Detector & SDXL & SD3.5 & Flux1 & Average \\
\toprule
 & $\mathrm{R_m}8$/$\mathrm{R_m}16$ & $\mathrm{R_m}8$/$\mathrm{R_m}16$ & $\mathrm{R_m}8$/$\mathrm{R_m}16$ & $\mathrm{R_m}8$/$\mathrm{R_m}16$ \\
\midrule
\cite{corvi2023detection} & $-0.18$/$\mathbf{-0.37}$ & $-0.21$/$\mathbf{-0.32}$ & $\mathbf{-0.49}$/$\mathbf{-0.54}$ & $-0.29$/$\mathbf{-0.41}$ \\
\cite{cozzolino2024raising} & $\mathbf{+1.27}$/$\mathbf{+1.69}$ & $\mathbf{+1.41}$/$\mathbf{+1.90}$ & $\mathbf{+1.42}$/$\mathbf{+1.83}$ & $\mathbf{+1.36}$/$\mathbf{+1.81}$ \\
\cite{mandelli2024synthetic} & $+0.00$/$+0.00$ & $-0.14$/$\mathbf{-0.32}$ & $-0.15$/$-0.23$ & $-0.10$/$-0.18$ \\
\cite{guillaro2025bias} & $-0.02$/$-0.04$ & $\mathbf{+0.32}$/$\mathbf{+0.37}$ & $\mathbf{+0.59}$/$\mathbf{+0.96}$ & $+0.30$/$\mathbf{+0.43}$ \\
\cite{baraldi2024contrasting} & $+0.02$/$-0.14$ & $+0.02$/$+0.11$ & $+0.00$/$+0.10$ & $+0.01$/$+0.02$ \\
\cite{sha2023fake} &$+0.02$/$-0.14$ & $+0.02$/$+0.11$ & $+0.00$/$+0.10$ & $+0.01$/$+0.02$ \\
\cite{koutlis24} & $\mathbf{+0.43}$/$+0.25$ & $+0.18$/$+0.11$ & $\mathbf{+0.42}$/$+0.14$ & $\mathbf{+0.34}$/$+0.17$ \\
Linear-$8$ & $\mathbf{-0.61}$/$\mathbf{-0.65}$ & $\mathbf{-0.77}$/$\mathbf{-0.85}$ & $\mathbf{-0.44}$/$\mathbf{-0.47}$ & $\mathbf{-0.61}$/$\mathbf{-0.66}$ \\
Linear-$16$& $\mathbf{-0.61}$/$\mathbf{-0.79}$ & $\mathbf{-0.77}$/$\mathbf{-0.93}$ & $\mathbf{-0.46}$/$\mathbf{-0.75}$ & $\mathbf{-0.61}$/$\mathbf{-0.82}$ \\
\bottomrule
\end{tabular}
}
\vspace{-15pt}
\end{table}

\textbf{Peaks removal from synthetic images. }We start evaluating detectors on the ``untouched'' synthetic data (i.e., without peaks removal), reporting results in Table~\ref{tab:wild_results}. All detectors achieve satisfactory performance across all generators, except for detectors~\cite{cozzolino2024raising, baraldi2024contrasting} which return an average \gls{tpr0} below $70\%$. Thus, we focus our further analysis on the remaining detectors. 
Notably, the linear detector achieves remarkably high performance on several generation techniques. While it would likely be easily fooled by simple JPEG compression (as noted earlier), it is striking that averaging contributions at specific frequencies alone produces such strong detection results.

Table~\ref{tab:rel_diff_rm_wild_agg} reports the relative differences between the \gls{tpr0} obtained under the peaks removal scenario and that of the standard case. To avoid confusion over deep learning-based detectors, we exclude datasets where the initial \gls{tpr0} (i.e., without peaks removal) was below $70\%$, thereby retaining only scenarios in which these detectors already demonstrated good detection performance.

Interestingly, among the deep learning-based detectors, only the one proposed in~\cite{corvi2023detection} appear to be strongly affected by the peaks removal operation, showing an average drop larger than $45\%$ in \gls{tpr0}. Moreover, this effect is not consistent across datasets, as this detector is not affected on DALL$\cdot$E\,3, Leonardo AI, and Starry AI images. A similar dataset-dependent behaviour is observed for other detectors, such as~\cite{guillaro2025bias, mandelli2024synthetic}, while some detectors like~\cite{sha2023fake, koutlis24} are not affected at all by the peaks removal procedure.

As expected, the linear detectors are strongly impaired by peaks removal, with two notable exceptions: Flux 1.1Pro and \gls{sd}3.5 generators. For these images, performance consistently drops only with the linear-$16$ detector under the ``$\mathrm{R_m}16$'' scenario. We hypothesize these generators carry significant energy contributions at periodicity $16$, thus removing peaks with grid $8$ is insufficient to degrade performance. This is consistent with the spectra reported in Fig.~\ref{fig:frequency_artifacts_wild}, which clearly show the $16$-step periodicity of both generators.

\textbf{Peaks removal from laundered images. }For brevity's sake, we omit the results obtained on real and laundered images without peak removal, though full details are available in the paper repository\footref{footnote_ref}. Table~\ref{tab:tpr_laundered_rel_diff} reports the relative difference in \gls{tpr0} between the peak removal scenario and the standard case. Interestingly, detector~\cite{corvi2023detection} shows a drop comparable to that observed on synthetic data, while detectors~\cite{guillaro2025bias, cozzolino2024raising, koutlis24} even show an increase in \gls{tpr0}. As expected, the linear detectors are far more interpretable, as both exhibit a performance drop consistent with the results previously observed.

\textbf{Results' discussion. }Overall, the results suggest that deep learning–based detectors do not exhibit a behaviour directly tied to the presence or absence of spectral peaks. The only detector showing an average behaviour that may be linked to these peaks is the one proposed in~\cite{corvi2023detection}. Indeed, its performance resembles that of linear detectors which, by design, must experience a performance drop when the spectral energy at the peak positions is reduced to zero.

By contrast, most of the other detectors appear largely unaffected by the presence of peaks. For instance, the detectors proposed in~\cite{sha2023fake, mandelli2024synthetic, baraldi2024contrasting} do not seem to rely on this information for detection. 
While it would be premature to conclude that frequency peaks are irrelevant for detection, this study represents a first step toward a clearer interpretability of results, questioning the assumption (often implicit in prior work) that such artifacts are necessarily exploited by deep learning algorithms, which are inherently black-box models.

One possible explanation for the apparent independence of these detectors from spectral peak energy is that they have been trained to be robust against data compression, which, as previously discussed, is known to introduce a periodicity in the frequency domain. Extensive data augmentation during training may encourage detectors to disregard such artifacts, as these could be easily masked by simple processing operations.

In any case, we find it striking that a simple linear detector can achieve extremely high accuracy. While its performance would inevitably drop under post-processing, we believe it is worth exploring future research on hybrid detectors that can combine the representational power of deep learning with the interpretability of model-based methods.

\section{Conclusions}
\label{sec:conclusions}

This work provides new insights into the role of frequency-domain artifacts, particularly spectral peaks, in synthetic image detection. Through systematic removal of these peaks and evaluation across multiple detectors, we find that most state-of-the-art deep learning detectors do not seem to significantly rely on them, challenging a common assumption in the field (frequently implicit in previous studies) that forensic detectors inevitably rely on such artifacts.

At the same time, the impressive performance of a simple linear peak-based detector highlights the potential of interpretable, model-based approaches, paving the way for possible hybrid strategies that can combine the transparency of linear methods with the representational power of deep learning.

\vfill\pagebreak

\bibliographystyle{IEEEbib}
\bibliography{strings,refs}

\begin{thebibliography}{10}

\bibitem{yang2024pixel}
Tao Yang, Rongyuan Wu, Peiran Ren, Xuansong Xie, and Lei Zhang,
\newblock ``Pixel-aware stable diffusion for realistic image super-resolution
  and personalized stylization,''
\newblock in {\em European conference on computer vision}. Springer, 2024, pp.
  74--91.

\bibitem{amerini2025deepfake}
Irene Amerini, Mauro Barni, Sebastiano Battiato, Paolo Bestagini, Giulia Boato,
  Vittoria Bruni, Roberto Caldelli, Francesco De~Natale, Rocco De~Nicola, Luca
  Guarnera, et~al.,
\newblock ``Deepfake media forensics: Status and future challenges,''
\newblock {\em Journal of Imaging}, vol. 11, no. 3, pp. 73, 2025.

\bibitem{corvi2023detection}
Riccardo Corvi, Davide Cozzolino, Giada Zingarini, Giovanni Poggi, Koki Nagano,
  and Luisa Verdoliva,
\newblock ``On the detection of synthetic images generated by diffusion
  models,''
\newblock in {\em ICASSP 2023-2023 IEEE International Conference on Acoustics,
  Speech and Signal Processing (ICASSP)}. IEEE, 2023, pp. 1--5.

\bibitem{guillaro2025bias}
Fabrizio Guillaro, Giada Zingarini, Ben Usman, Avneesh Sud, Davide Cozzolino,
  and Luisa Verdoliva,
\newblock ``A bias-free training paradigm for more general ai-generated image
  detection,''
\newblock in {\em Proceedings of the Computer Vision and Pattern Recognition
  Conference}, 2025, pp. 18685--18694.

\bibitem{cozzolino2024raising}
Davide Cozzolino, Giovanni Poggi, Riccardo Corvi, Matthias Nie{\ss}ner, and
  Luisa Verdoliva,
\newblock ``Raising the bar of ai-generated image detection with clip,''
\newblock in {\em Proceedings of the IEEE/CVF Conference on Computer Vision and
  Pattern Recognition}, 2024, pp. 4356--4366.

\bibitem{sha2023fake}
Zeyang Sha, Zheng Li, Ning Yu, and Yang Zhang,
\newblock ``De-fake: Detection and attribution of fake images generated by
  text-to-image generation models,''
\newblock in {\em Proceedings of the 2023 ACM SIGSAC conference on computer and
  communications security}, 2023, pp. 3418--3432.

\bibitem{mandelli2024synthetic}
Sara Mandelli, Paolo Bestagini, and Stefano Tubaro,
\newblock ``When synthetic traces hide real content: Analysis of stable
  diffusion image laundering,''
\newblock in {\em 2024 IEEE International Workshop on Information Forensics and
  Security (WIFS)}. IEEE, 2024, pp. 1--6.

\bibitem{baraldi2024contrasting}
Lorenzo Baraldi, Federico Cocchi, Marcella Cornia, Lorenzo Baraldi, Alessandro
  Nicolosi, and Rita Cucchiara,
\newblock ``Contrasting deepfakes diffusion via contrastive learning and
  global-local similarities,''
\newblock in {\em European Conference on Computer Vision}. Springer, 2024, pp.
  199--216.

\bibitem{koutlis24}
Christos Koutlis and Symeon Papadopoulos,
\newblock ``Leveraging representations from intermediate encoder-blocks for
  synthetic image detection,''
\newblock in {\em European conference on computer vision}. 2024, vol. 15130,
  pp. 394--411, Springer.

\bibitem{Corvi2023Intriguing}
Riccardo Corvi, D.~Cozzolino, G.~Poggi, Koki Nagano, and L.~Verdoliva,
\newblock ``Intriguing properties of synthetic images: from generative
  adversarial networks to diffusion models,''
\newblock {\em 2023 IEEE/CVF Conference on Computer Vision and Pattern
  Recognition Workshops (CVPRW)}, pp. 973--982, 2023.

\bibitem{bammey2023synthbuster}
Quentin Bammey,
\newblock ``Synthbuster: Towards detection of diffusion model generated
  images,''
\newblock {\em IEEE Open Journal of Signal Processing}, vol. 5, pp. 1--9, 2023.

\bibitem{coccomini2024deepfake}
Davide~Alessandro Coccomini, Roberto Caldelli, Claudio Gennaro, Giuseppe
  Fiameni, Giuseppe Amato, and Fabrizio Falchi,
\newblock ``Deepfake detection without deepfakes: Generalization via synthetic
  frequency patterns injection,''
\newblock {\em arXiv preprint arXiv:2403.13479}, 2024.

\bibitem{durall2020watch}
Ricard Durall, Margret Keuper, and Janis Keuper,
\newblock ``Watch your up-convolution: Cnn based generative deep neural
  networks are failing to reproduce spectral distributions,''
\newblock in {\em Proceedings of the IEEE/CVF conference on computer vision and
  pattern recognition}, 2020, pp. 7890--7899.

\bibitem{zhang2017beyond}
Kai Zhang, Wangmeng Zuo, Yunjin Chen, Deyu Meng, and Lei Zhang,
\newblock ``Beyond a gaussian denoiser: Residual learning of deep cnn for image
  denoising,''
\newblock {\em IEEE transactions on image processing}, vol. 26, no. 7, pp.
  3142--3155, 2017.

\bibitem{bongini2025wild}
Pietro Bongini, Sara Mandelli, Andrea Montibeller, Mirko Casu, Orazio Pontorno,
  Claudio~Vittorio Ragaglia, Luca Zanchetta, Mattia Aquilina, Taiba~Majid Wani,
  Luca Guarnera, et~al.,
\newblock ``Wild: a new in-the-wild image linkage dataset for synthetic image
  attribution,''
\newblock {\em arXiv preprint arXiv:2504.19595}, 2025.

\bibitem{stable_diffusion_35}
Stability AI,
\newblock ``Stable diffusion 3.5-large,''
  \url{https://huggingface.co/stabilityai/stable-diffusion-3.5-large}.

\bibitem{fluxpro}
{Black Forest Labs},
\newblock ``{FLUX 1.1 [pro]: Advanced Text-to-Image Generation Model},'' 2024.

\bibitem{dalle3}
OpenAI,
\newblock ``Improving image generation with better captions,''
  \url{https://cdn.openai.com/papers/dall-e-3.pdf}, 2024.

\bibitem{cannas2024jpeg}
Edoardo~Daniele Cannas, Sara Mandelli, Nata{\v{s}}a Popovi{\'c}, Ayman
  Alkhateeb, Alessandro Gnutti, Paolo Bestagini, and Stefano Tubaro,
\newblock ``Is jpeg ai going to change image forensics?,''
\newblock {\em arXiv preprint arXiv:2412.03261}, 2024.

\bibitem{zhang2019detecting}
Xu~Zhang, Svebor Karaman, and Shih-Fu Chang,
\newblock ``Detecting and simulating artifacts in gan fake images,''
\newblock in {\em 2019 IEEE international workshop on information forensics and
  security (WIFS)}. IEEE, 2019, pp. 1--6.

\bibitem{dzanic2020fourier}
Tarik Dzanic, Karan Shah, and Freddie Witherden,
\newblock ``Fourier spectrum discrepancies in deep network generated images,''
\newblock {\em Advances in neural information processing systems}, vol. 33, pp.
  3022--3032, 2020.

\bibitem{frank2020leveraging}
Joel Frank, Thorsten Eisenhofer, Lea Sch{\"o}nherr, Asja Fischer, Dorothea
  Kolossa, and Thorsten Holz,
\newblock ``Leveraging frequency analysis for deep fake image recognition,''
\newblock in {\em International conference on machine learning}. PMLR, 2020,
  pp. 3247--3258.

\bibitem{midjourney}
{MidJourney},
\newblock ``{MidJourney: An AI-powered image generation tool},'' 2024.

\bibitem{podell2023sdxl}
Dustin Podell, Zion English, Kyle Lacey, Andreas Blattmann, Tim Dockhorn, Jonas
  M{\"u}ller, Joe Penna, and Robin Rombach,
\newblock ``Sdxl: Improving latent diffusion models for high-resolution image
  synthesis,''
\newblock {\em arXiv preprint arXiv:2307.01952}, 2023.

\bibitem{firefly}
Adobe,
\newblock ``Adobe firefly,'' \url{{https://firefly.adobe.com/}}, 2023.

\bibitem{freepik}
{Freepik},
\newblock ``{Freepik AI Image Generator},''
  \url{https://docs.freepik.com/api-reference/mystic/post-mystic}, 2024.

\bibitem{dang2015raise}
Duc-Tien Dang-Nguyen, Cecilia Pasquini, Valentina Conotter, and Giulia Boato,
\newblock ``Raise: A raw images dataset for digital image forensics,''
\newblock in {\em Proceedings of the 6th ACM multimedia systems conference},
  2015, pp. 219--224.

\bibitem{flux2024}
Black~Forest Labs,
\newblock ``Flux,'' \url{https://github.com/black-forest-labs/flux}, 2024.

\bibitem{marr1980theory}
David Marr and Ellen Hildreth,
\newblock ``Theory of edge detection,''
\newblock {\em Proceedings of the Royal Society of London. Series B. Biological
  Sciences}, vol. 207, no. 1167, pp. 187--217, 1980.

\end{thebibliography}

\end{document}